\documentclass[review]{elsarticle}

\usepackage[utf8]{inputenc}
\usepackage[T1]{fontenc}
\usepackage{hyperref}
\usepackage{url}
\usepackage{amsfonts}
\usepackage{nicefrac}
\usepackage{microtype}
\usepackage{graphicx}
\usepackage{float}
\usepackage[justification=centering]{caption}
\usepackage{todonotes}
\usepackage{natbib}
\biboptions{authoryear}
\usepackage{booktabs}
\usepackage{multirow}

\begin{document}

\begin{frontmatter}

\title{Interpretable Predictive Maintenance for Hard Drives}

\author[iai]{Maxime Amram}
\ead{maxime@interpretable.ai}
\author[iai]{Jack Dunn\corref{mycorrespondingauthor}}
\cortext[mycorrespondingauthor]{Corresponding author}
\ead{jack@interpretable.ai}
\author[iai]{Jeremy J. Toledano}
\ead{jeremy@interpretable.ai}
\author[iai]{Ying Daisy Zhuo}
\ead{daisy@interpretable.ai}

\address[iai]{Interpretable AI, Cambridge, MA 02142, USA}

\begin{abstract}
Existing machine learning approaches for data-driven predictive maintenance are usually black boxes that claim high predictive power yet cannot be understood by humans. This limits the ability of humans to use these models to derive insights and understanding of the underlying failure mechanisms, and also limits the degree of confidence that can be placed in such a system to perform well on future data. We consider the task of predicting hard drive failure in a data center using recent algorithms for interpretable machine learning. We demonstrate that these methods provide meaningful insights about short- and long-term drive health, while also maintaining high predictive performance. We also show that these analyses still deliver useful insights even when limited historical data is available, enabling their use in situations where data collection has only recently begun.
\end{abstract}


\end{frontmatter}

\section{Introduction}

Backblaze is a large data storage service provider with over 130,000 hard drives in its data centers. Hard drives have a number of sensors that are continuously monitoring and reporting on the health of the drive, known as SMART (Self-Monitoring, Analysis and Reporting Technology). Backblaze regularly publishes the historical daily snapshots of the SMART statistics for each of its hard drives~\citep{backblaze}. Additionally, they record whenever a hard drive fails or is removed from service.

The ability to understand and predict when a hard drive is going to fail is extremely valuable to the operations of such a data center, as it offers the ability to avoid downtime through better timing for replacement and repair of equipment (preventive maintenance), as well as providing more input into strategic planning and forecasts of operational needs. Based on univariate analysis and domain knowledge,~\cite{backblazeblog} identified five SMART metrics (numbers 5, 187, 188, 197, and 198) that it uses as indicators of impending failure (these findings will be presented in more detail in Section \ref{overview}.)

The use of machine learning methods for such predictive maintenance tasks is growing in popularity, particularly with the rapid increase in deployment of IoT (internet of things) devices and sensors. While humans have trouble considering more than a few variables at a time, machine learning and artificial intelligence methods are capable of handling thousands of variables at a time when building predictive models. These approaches are more powerful than simple univariate analyses as they can model complex interaction effects between features rather than simply considering each variable in isolation.

Despite their increased power over simpler approaches, these machine learning methods are often ``black-box'' approaches whose prediction logic is difficult or impossible for humans to comprehend. This is undesirable, as although we may be able to build a predictive model that can predict drive failure on historical data with high accuracy, it is difficult to extract insights about how the features interact to indicate impending failure, and also hard to have full confidence that such a system will continue to perform well if deployed in production.

An active area of research is interpretable machine learning methods, which unlike black-box methods are completely understandable by humans. Recent work has demonstrated that modern optimization techniques can be leveraged to construct interpretable methods with performance rivaling the black-box methods, enabling us to leverage the insights and confidence that interpretability brings without sacrificing performance~\citep{bertsimas2019machine}.

The main goal of our analysis is to develop a data-driven model for understanding how the SMART metrics relate to impending failure. Given the desire to learn and understand how these metrics can indicate failure, it is natural that we consider approaching this task with interpretable models. In particular, in this paper we concern ourselves with understanding the following two dimensions of hard drive health:
\begin{itemize}
    \item the long-term health of a hard drive over its expected lifespan (roughly 3 years)
    \item the short-term health of a hard drive and whether it is likely to fail in the immediate future (30 to 90 days)
\end{itemize}

We also consider these problems from the perspective of limited data. In the case of Backblaze, there are many years of rich data available that we can use for modeling. However, we would like to understand whether we can use data from a limited timespan to conduct the same analysis and arrive at similar conclusions. Such a result would prove useful in situations where sensors have only recently been installed, permitting this kind of analysis without waiting years to collect sufficient data.

We summarize our contributions in this paper:
\begin{itemize}
    \item Using data across a three-year time horizon, we build an Optimal Survival Tree (OST) to model the long-term health of drives. We observe that the resulting model places importance on metrics known to be correlated with drive failure, as well as additional features not previously highlighted. Moreover, the model demonstrates that these features interact in specific ways when indicating the outlook for drives, where some metrics are only useful for healthy drives, while others are relevant when examining drives with high risk of imminent failure.
    \item Using data from a shorter time horizon, we build both OST and Optimal Classification Tree (OCT) models to predict the risk of short-term failure (30--90 days). These models again surface feature interactions that are highly predictive of drive failure, and both have strong predictive performance (sensitivity approximately 50\% with a false alarm rate around 10\%).
    \item Finally, we demonstrate that these analyses can be applied to deliver similar insights even in scenarios where a wealth of historical data is not available. This is particularly significant as it is often the case that sensors have only recently been installed, limiting how much data is available. We show that even in this data-poor case, our approach can still be leveraged to deliver useful insights.
\end{itemize}

The structure of the paper is as follows. In Section \ref{literature}, we review the classical approaches used in the literature for these problems. In Section \ref{overview}, we give an overview of the data published by Backblaze, as well as their existing findings. In Section \ref{long-term}, we consider the problem of predicting long-term health using OST. In Section \ref{short-term}, we use OCT and OST to predict short-term failure. In Section \ref{limited-data}, we repeat our analysis using data from a limited time frame to simulate the case where limited historical data is available. Finally, in Section \ref{conclusion} we present a summary of our findings.

\subsection*{Literature review} \label{literature}

Machine learning has become a popular approach for solving predictive maintenance problems in the literature (see~\cite{carvalho2019systematic} for a comprehensive review). When data on the occurrence of failures is available, \emph{supervised} machine learning is typically the preferred approach, as they tend to be more powerful than \emph{unsupervised} approaches, which learn purely from the sensor data and do not require failure labels~\citep{kanawaday2017machine}. In some cases, the occurrence of failure may not be recorded, or failures may not be observed (e.g. if machines are rotated out of service before failure or preemptively maintained), necessitating the use of unsupervised methods. In our case, failure data is available so we focus on supervised approaches.

Supervised machine learning uses data comprised of pairs of the form $(\mathbf{x}_i, y_i), i = 1, \ldots, n$. Each $\mathbf{x}_i$ is a $p$-dimensional vector containing the measurements for each sensor (in our case, the daily SMART values), and $y_i$ represents the outcome variable. Depending on the choice of outcome variable, we can generate different classes of problem:
\begin{itemize}
  \item $y_i$ representing a binary or categorical outcome (e.g. whether the drive failed within 30 days) gives a \emph{classification} problem
  \item $y_i$ representing a numerical outcome (e.g. the time until the drive failed) gives a \emph{regression} or \emph{survival} problem
\end{itemize}

Many different algorithms can be used to solve a classification problem, classical approaches include both interpretable methods like logistic regression and decision trees (CART), as well as black-box methods like random forests and gradient boosting~\citep{friedman2001elements}. Traditionally, black-box methods have outperformed interpretable methods in terms of predictive performance, resulting in a significant price of interpretability and a difficult tradeoff for practitioners. Recent works combining machine learning with modern mathematical optimization have developed Optimal Classification Trees~\citep{bertsimas2017optimal,bertsimas2019machine} which permits constructing a single decision tree that has similar performance to the black-box methods, thus delivering interpretability without sacrificing performance.

When it comes to predicting continuous outcomes such as the remaining useful life, a standard regression analysis is often not well-suited to solving the problem. This is because it is common that the majority of the records in the data have not experienced a failure, and thus we have no measure of their remaining useful life. However, we can still use these data for learning by exploiting the fact that we have observed these machines without failure for some time, and using this to form lower bounds for remaining lifespan. For example, if we have observed a machine without failure for 90 days, then we know that as of the first day, it has a remaining useful life of \emph{at least} 90 days. Survival analysis is a specialized class of models designed to deal with such so-called \emph{censored} data. A classical survival analysis tool is the Cox proportional hazards model, which models impact of features on the lifespan using an approach analogous to linear regression~\citep{cox1972regression}. Another traditional approach is the Kaplan-Meier estimator, which is a non-parametric model of survival over time~\citep{kaplan1958nonparametric}. Decision tree models can also be applied to survival problems for enhanced power and interpretability. Just as Optimal Classification Trees improve significantly over traditional tree methods for classification problems, Optimal Survival Trees~\citep{bertsimas2020survival} use modern optimization approaches to improve significantly over existing greedy survival tree methods~\citep{leblanc1992relative} while maintaining interpretability.

\section{Overview of the Backblaze data and insights} \label{overview}
In this section, we present an overview of the SMART data collected and published by Backblaze. We also review their existing research and efforts to link SMART metrics to hard drive failure.

\subsection{Backblaze data}

Each of Backblaze's 130,000 hard drives is monitored daily and its activity is recorded as a single row in a daily data file. Every day, for each hard drive, Backblaze records several daily aggregated SMART metrics as well as whether it failed on this day. Each attribute measures different characteristics about the drive's operation, and we present a summary of the most relevant quantities in Table~\ref{tab:smart}.

\begin{figure}
  \centering
  \includegraphics[scale=0.5]{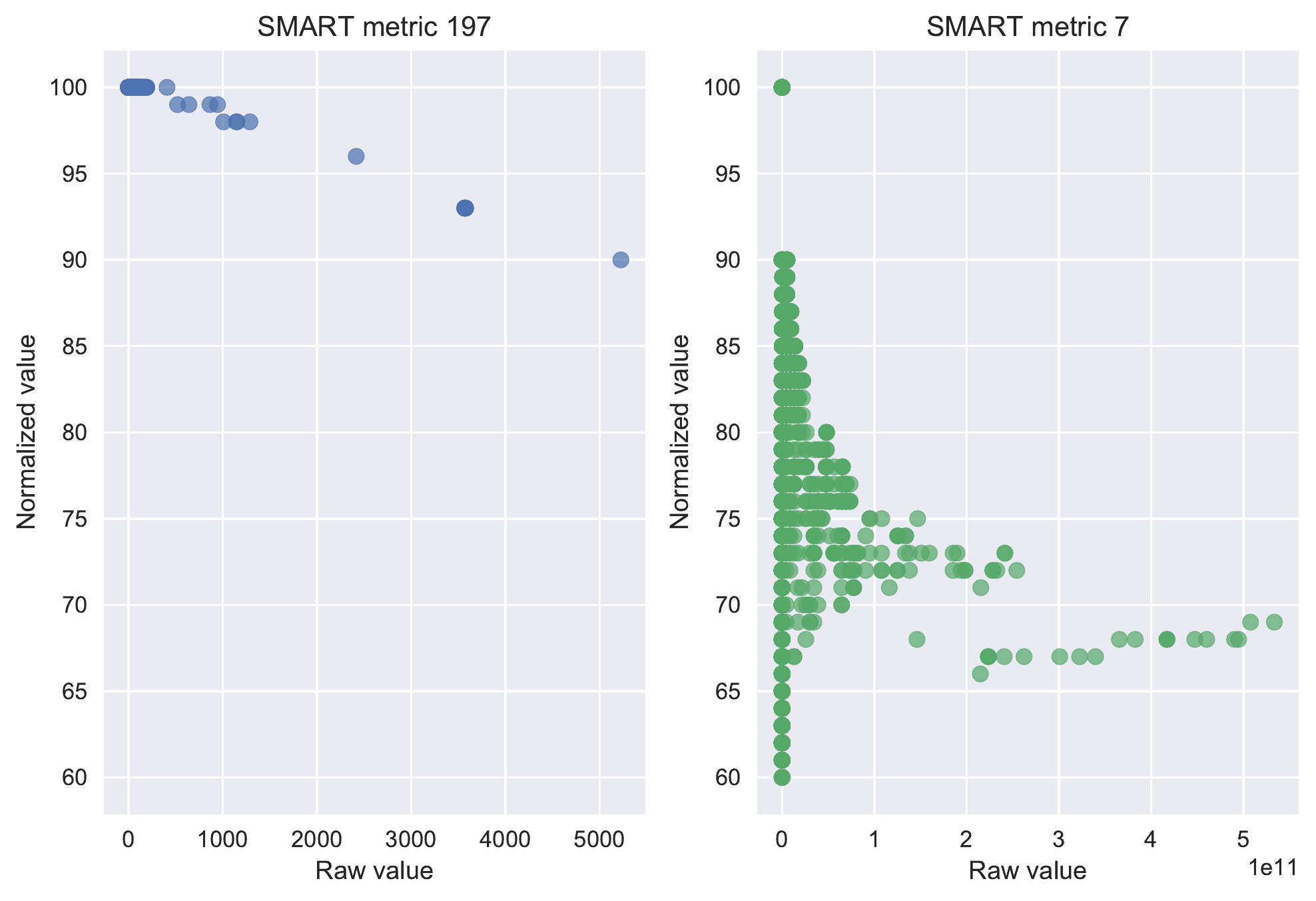}
  \caption{Normalized values as function of the raw values for metrics 197 and 7}
  \label{fig:raw-norm}
\end{figure}

Each SMART attribute has a raw value whose meaning is determined entirely by the drive manufacturer (but often corresponds to the raw physical unit that is being measured), as well as a normalized value between 1 and 253, where 1 is worst, and 100 is usually the starting value~\citep{wikismart}.  Figure~\ref{fig:raw-norm} shows the mapping between the normalized and the raw measurements for SMART metrics 197 and 7. We see that for metric 197 the raw values map directly to the normalized values, whereas for metric 7 there is little correlation between the raw and normalized metrics. Although most metrics in the dataset show a similar pattern to 197 with a clear mapping between the values, there are many metrics that, like metric 7, show little correlation between the values. As a result, in our analysis we will opt to retain both raw and normalized values in case having these different signals results in more predictive power.

\renewcommand{\arraystretch}{1.5}
\begin{table}
    \centering
    \scriptsize
    \begin{tabular}{clp{5cm}}
        \toprule
        \multicolumn{1}{c}{\textbf{SMART}} &
          \multicolumn{1}{c}{\textbf{Attribute Name}} &
          \multicolumn{1}{c}{\textbf{Description}} \\ \midrule
        \textbf{3} &
          Spin-Up Time &
          Average time (in milliseconds) of spindle spin up from zero RPM to fully operational. \\
        \textbf{5} &
          Reallocated Sectors Count &
          Count of bad sectors that have been found and reallocated. A hard drive which has had a reallocation is very likely to fail in the immediate months. \\
        \textbf{7} &
          Seek Error Rate &
          Rate of seek errors of the magnetic heads, due to partial failure in the mechanical positioning system. \\
        \textbf{187} &
          Reported Uncorrectable Errors &
          Count of errors that could not be recovered using hardware ECC (Error-Correcting Code), a type of memory used to correct data corruption errors. \\
        \textbf{188} &
          Command Timeout &
          Count of aborted operations due to hard drive timeout. \\
        \textbf{190} &
          Temperature Difference &
          Difference between current hard drive temperature and optimal temperature of 100$^{\circ}$C. \\
        \textbf{197} &
          Current Pending Sectors Count &
          Count of bad sectors that have been found and waiting to be reallocated, because of unrecoverable read errors. \\
         \textbf{198} &
          Offline Uncorrectable Sectors Count &
          Total count of uncorrectable errors when reading/writing a sector, indicating defects of the disk surface or problems in the mechanical subsystem. \\ \bottomrule
    \end{tabular}%

    \caption{Descriptions of relevant SMART metrics}\label{tab:smart}
\end{table}

At the end of Q1 2020, Backblaze had recorded a total of 7,033 hard drive failures since April 2013 for the 130,000 hard drives still in service in 2020, for an annualized failure rate of 1.71\%~\citep{backblaze2020}.

\begin{figure}
  \centering
  \includegraphics[scale=0.45]{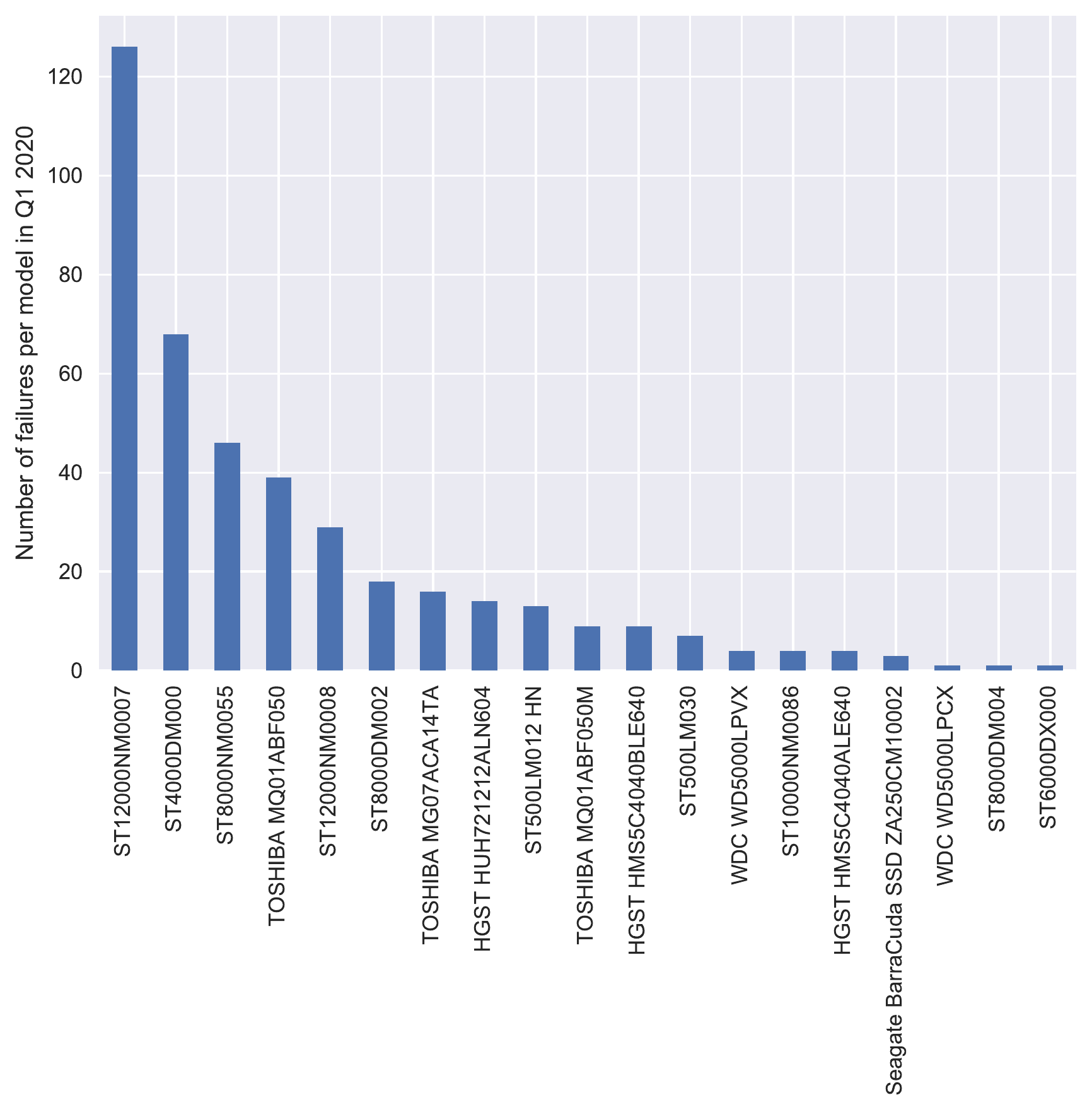}
  \caption{Number of failures per model in Q1 2020}
  \label{fig:model-failures}
\end{figure}

Backblaze notes that there is significant variation in failure rates across different models of hard drive. Coupled with the potential for inconsistent meanings across models and manufacturers, we chose to limit our analysis to a single model of hard drive. Figure~\ref{fig:model-failures} shows the number of failures per model for Q1 2020. We see that model ST12000NM0007 accounts for roughly 30\% of the overall failures. Throughout the paper, we will thus focus our analysis on this specific hard drive model to maximize the amount of available data.

Hard drives are organized into ``storage pods'', which are distinct servers each with 60 hard drives. Backblaze labels a hard drive as failing whenever it is removed from its storage pod~\citep{backblazeblog}. There are two possible reasons for such a removal:
\begin{enumerate}
    \item The hard drive has stopped working: it is impossible to turn it on, it does not respond to console commands, or it cannot be read or written anymore.
    \item The hard drive has been deemed ``about to fail'' (indicated by a positive value for SMART metric 187, which means that the hard drive has started experiencing uncorrectable errors).
\end{enumerate}
It is necessary for Backblaze to anticipate such failures as a single failing hard drive could impede the whole storage pod's operation.

\subsection{Existing analysis of SMART data}

\subsubsection*{Backblaze}

In an effort to understand how SMART statistics relate to drive failure, \cite{backblazeblog} ran a simple univariate correlation analysis between each SMART metric and their failure records. Based on this analysis, coupled with their own domain knowledge, they concluded that the following SMART metrics are good predictors of impending failure:
\begin{itemize}
    \item 5 (Reallocated Sectors Count)
    \item 187 (Reported Uncorrectable Errors)
    \item 188 (Command Timeout)
    \item 197 (Current Pending Sectors Count)
    \item 198 (Offline Uncorrectable Sectors Count)
\end{itemize}
Additionally, they claim these metrics have the benefit of generally being consistent across hard drive manufacturers.

\subsubsection*{Google}

\cite{google} conducted a detailed correlation analysis of SMART metrics for a large number of hard drives in operation at Google. We summarize their findings as follows:
\begin{itemize}
    \item Four SMART metrics are consistently highly correlated with failure: 5, 187, 197 and 198.
    \item Six additional SMART metrics were found to be relevant in predicting failures, but are not always consistent across models and manufacturers, including 7 (Seek Errors).
    \item They find that any change in metric 187 is highly predictive of failure: ``after their first scan error (i.e. when a positive value for 187 is observed for the first time), drives are 39 times more likely to fail within 60 days than drives with no such errors.''
    \item Temperature, which is often considered the most important factor in predicting failures by many in the industry, was not found to be positively correlated with failure in relatively young hard drives.
\end{itemize}

\subsubsection*{Limitations}

Both of these analyses were univariate and only considered correlation between failures and a single metric at a time. As such, they would not be able to detect any nonlinear interactions between metrics that affected the chance of failure. Another limitation of this analysis is that it leaves humans to choose the cutoff values that will raise alerts if exceeded.

\section{Predicting long-term health} \label{long-term}

In this section, we address the problem of trying to predict the overall health of a hard drive over a long time horizon using the SMART metrics.

\subsection{Data Processing}

As stated earlier, our analysis is restricted to model ST12000NM0007. We isolated hard drives that failed between Q1 2019 and Q1 2020 (inclusive), and gathered their daily SMART metrics records for the past three years (back to Q1 2017). This data captured the entire lifespan for the majority of these drives, indicating we have data on their full trajectory since being put into service.

To keep our analysis interpretable and intuitive, we used only the raw and normalized SMART metrics captured in the data, and did not engineer any more complex features. Additionally, to allow our predictive models to focus on characterizing the health of the drive, we removed SMART metrics which represent cumulative counts over the lifespan of the drive and thus are highly correlated with time (4, 9, 12, 192, 193, 240, 241, and 242). This allowed us to remove from the model the obvious result that older drives are more likely to fail, and as a result we can uncover more interesting predictive insights based on metrics that are more related to the overall health status of the drives, regardless of age.

For each hard drive and for each daily snapshot, we compute the \textit{remaining useful life}, being the number of days until failure. Note that because we captured the complete lifespan for most of these drives, we have very little censored data.

\subsection{Models}

We used the daily SMART feature records coupled with remaining useful life values to train an Optimal Survival Tree that models how the remaining life is affected by SMART values. The resulting model is a single tree that uses the SMART metrics to segment the data into subpopulations with similar survival prognosis.

\begin{figure}
  \centering
  \includegraphics[width=\textwidth]{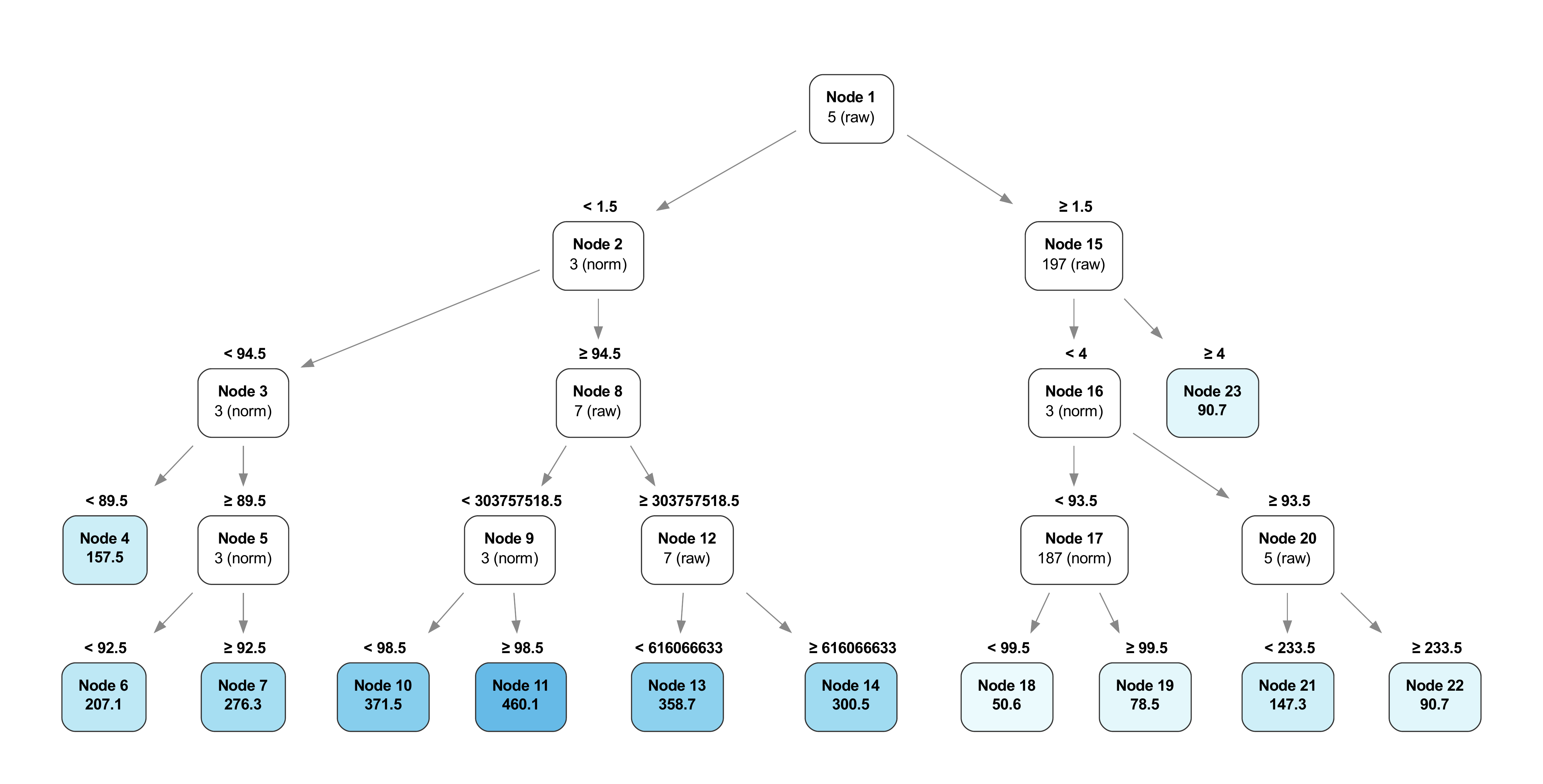}
  \caption{Optimal Survival Tree for predicting long-term health. The variable in the split nodes refer to the splitting SMART variable (either as raw or normalized value). The numbers in the leaf nodes refer to expected survival times in days, where darker shading in color corresponds to longer survival. }
  \label{fig:model-1-tree}
\end{figure}

\begin{figure}
  \centering
  \includegraphics[width=0.8\textwidth]{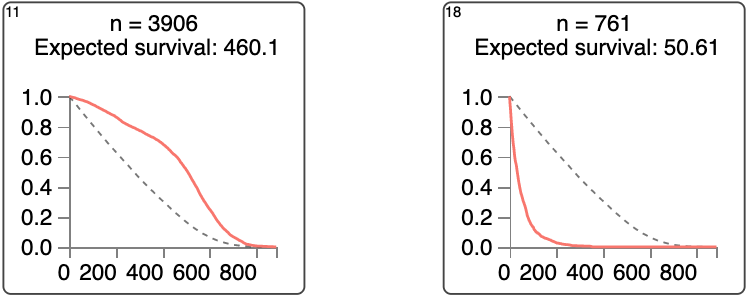}
  \caption{Survival curves for Nodes 11 and 18 in the Optimal Survival Tree for predicting long-term health. The x-axis is the time in days and the y-axis is the probability that a given drive will not fail before this time. The black dotted line in each curve is the overall survival curve for all hard drives, whereas the red line refers to the survival curve for hard drives in the specific leaf nodes.}
  \label{fig:model-1-curves}
\end{figure}

We show the trained survival tree in Figure \ref{fig:model-1-tree}. Starting from the top of the tree, the split nodes use one SMART feature at a time to recursively divide the hard drives. Each leaf node displays the expected survival time among hard drives in that leaf, with a darker color shading indicating longer survival. Each leaf node is also characterized by a Kaplan-Meier curve, showing the survival outlook over time for hard drives that fall into that leaf. Figure \ref{fig:model-1-curves} shows examples of these curves for two of the leaves in the tree (the full tree showing the Kaplan-Meier curves in each node is available in the supplementary materials).

We see that three out of the five SMART metrics identified by Backblaze as indicative of impending failure (namely 5, 187, and 197) are used by our model. In particular, the tree begins splitting on 5, indicating this is the most impactful metric to consider before following up with subsequent questions.

If metric 5 is below 1.5, we go to the left side of the tree, where survival tends to be higher, and subsequent splits are based on metrics 3 and 7. If on the other hand metric 5 is above 1.5, we go to the right side of the tree, with lower survival times that are refined by examining metrics 3, 5 (again), 187 and 197. This dichotomy shows the power of our advanced modeling over simple univariate analysis: we are able to identify cases where the features interact to affect survival differently, and some metrics only become relevant based on the values of others.

From the model we can also derive paths to failure that showcase how, under certain conditions (represented by SMART metrics values), failures are historically more likely to occur. In particular, we see that Node 18 is a good example of an extreme failure mode. This leaf node aggregates a population of machines, which, based on certain SMART metric conditions, have the lowest expected survival time of all of the leaf nodes displayed: after 50 days, most of the machines had failed, and all had failed within a year. The conditions chosen by the tree to assign a machine to Node 18 are symptomatic of an extreme failure mode, as such machines are experiencing:

\begin{itemize}
  \item High count of reallocations (bad sectors that have been found and reallocated), as shown by the positive value for raw SMART metric 5.
  \item High spin-up time (these machines are thus slower than usual, potentially indicating accumulated wear on the drive), as shown by a lower than 100 value for normalized SMART metric 3.
  \item Higher than usual count of uncorrectable errors, as shown by a lower than 100 value for normalized SMART metric 187.
\end{itemize}

This provides a very interpretable and understandable characterization of a set of hard drives that are in very poor health and should be expected to fail imminently.

The survival tree also uncovers subpopulations with characteristically healthy behaviors in hard drives. In particular, we consider Node 11, which has the highest expected survival time of all leaf nodes. The SMART metric conditions leading a hard drive to Node 11 also make intuitive sense: healthy values for raw SMART metrics 3, 5 and 7. Around half of the drives in this subpopulation end up lasting another two years, indicating they are indeed particularly healthy.

\begin{figure}
  \centering
  \includegraphics[scale=0.65]{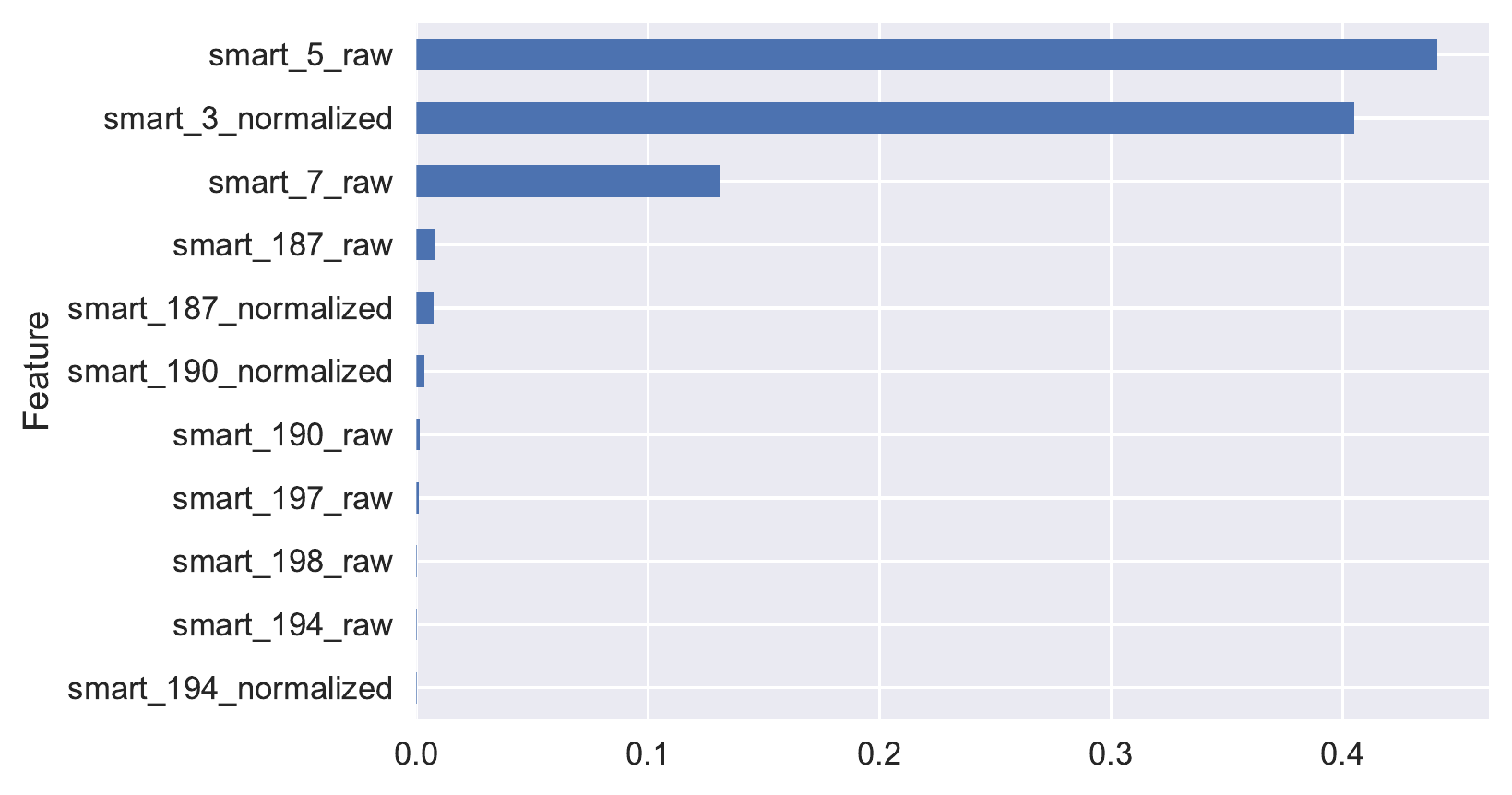}
  \caption{Variable importance for long-term Optimal Survival Tree}
  \label{fig:model-1-importance}
\end{figure}

Finally, to quantify the relative importance of the metrics in modeling long-term health, we can analyzing the variable importance plot for the model, shown in Figure \ref{fig:model-1-importance}. We see that metrics 3 and 5 are the most important determinants of the long-term health, followed by 7. The importance of 5 is particularly evident in the tree, as the life expectancy drops dramatically as soon as more than one bad sector has been found and reallocated.

\section{Predicting short-term health} \label{short-term}

In this section, we focus on the problem of predicting short-term drive failure from the SMART metrics. First, we formulate the problem as a short-term survival analysis and use Optimal Survival Trees to model drive health over this horizon, and then we pose the problem as a classification task and predict the occurence of failure within this time window.

\subsection{Short-term health as a survival problem} \label{ost}

For this analysis, we will consider only a single quarter's worth of data (Q1 2020), and use Optimal Survival Trees to model the drive health over this 90-day window.

Unlike Section \ref{long-term} where we limited the analysis to failing hard drives, in this analysis we will consider both hard drives which failed and hard drives which did not fail within the time window. This results in a large number of \textit{censored} observations (as discussed in Section~\ref{literature}), as most of the drives observed do not fail within the quarter. However, as discussed previously, the absence of failure of such drives can be utilized by survival models, because they still provide a lower bound on the remaining useful life of each drive.

\begin{figure}
  \centering
  \includegraphics[width=\textwidth]{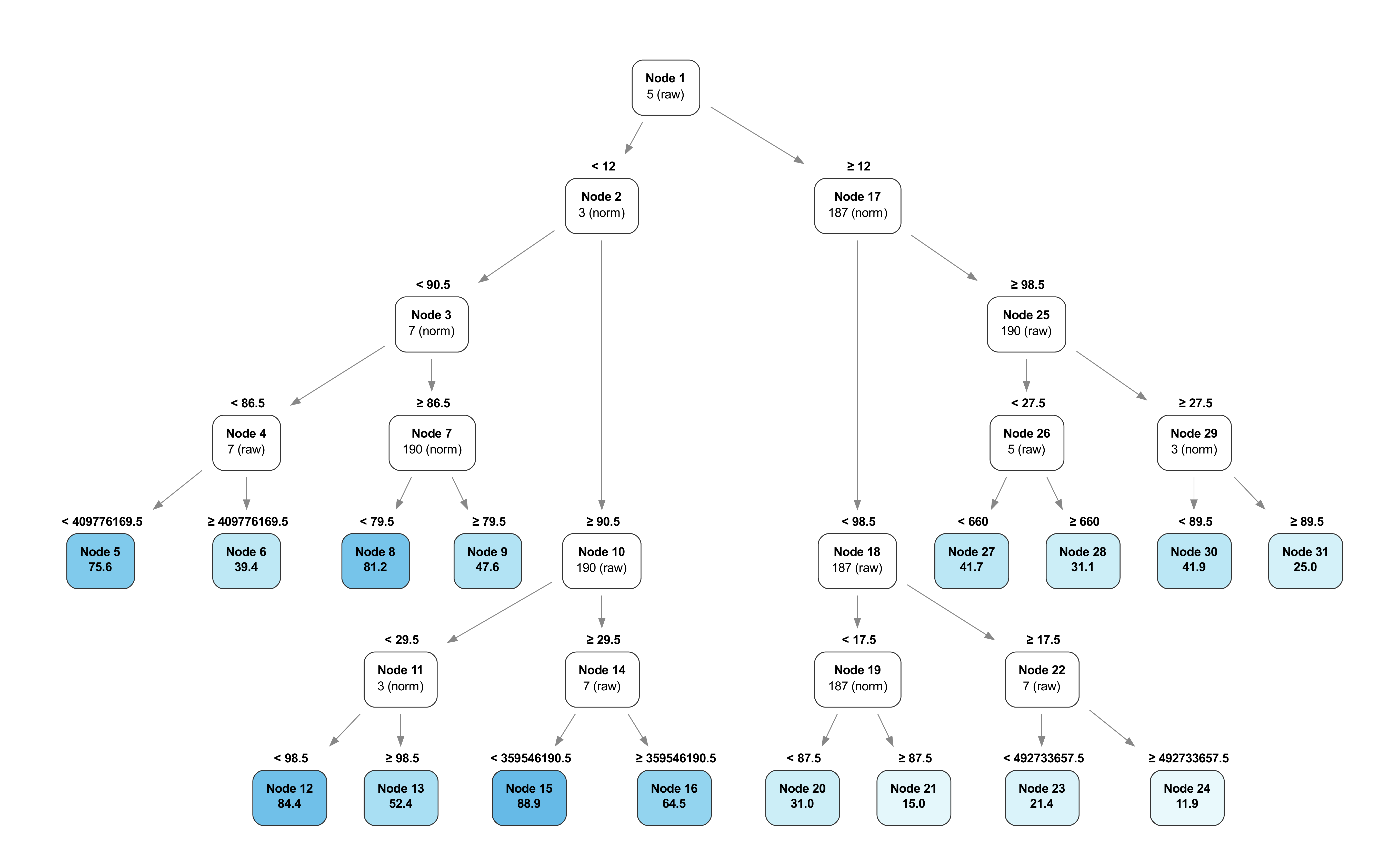}
  \caption{Optimal Survival Tree for predicting short-term health}
  \label{fig:model-4-tree}
\end{figure}

We used this dataset to train an Optimal Survival Tree, which is shown in Figure \ref{fig:model-4-tree} (the full interactive tree with Kaplan-Meier curves for each leaf is again available in the supplementary materials). As in Section \ref{long-term}, each leaf displays the expected survival time for drives falling into the leaf, with a darker color shading indicating longer survival.

We see that many of the features used by this survival tree are very similar to the ones used by the long-term health tree, and overlap heavily with the metrics identified by Backblaze and Google as highly correlated with failure.

\begin{figure}
  \centering
  \includegraphics[scale=0.65]{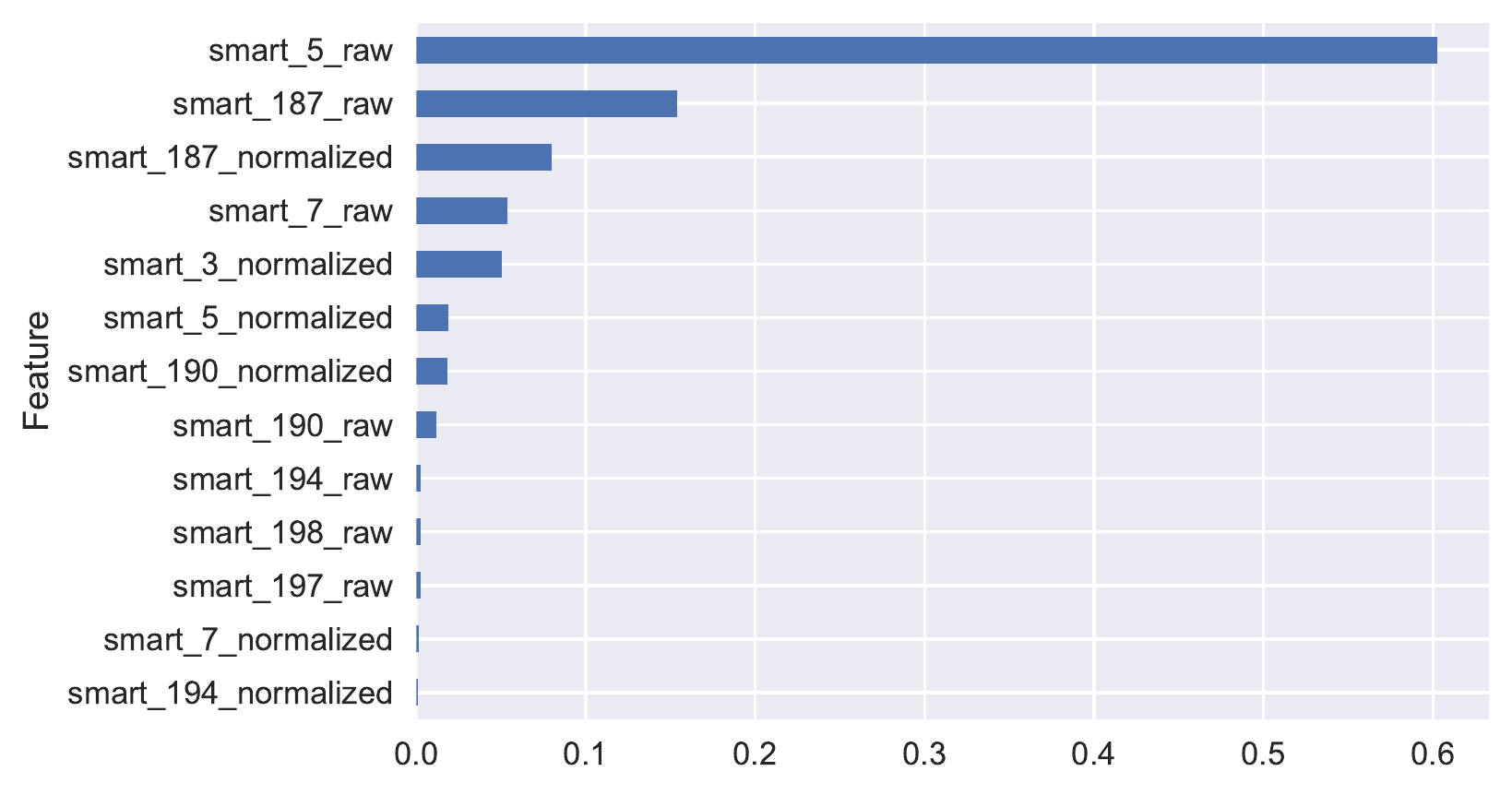}
  \caption{Variable importance for short-term Optimal Survival Tree}
  \label{fig:model-4-importance}
\end{figure}

The variable importance scores for this tree are shown in Figure \ref{fig:model-4-importance}. We see that metric 5 is again the most important, however metric 3 is significantly less important here compared to its importance in modeling long-term drive health. Instead, metric 187 becomes very important, mirroring the conclusion from Google that any change in this metric from baseline is a very significant predictor of failure.

\begin{figure}
  \centering
  \includegraphics[width=\textwidth]{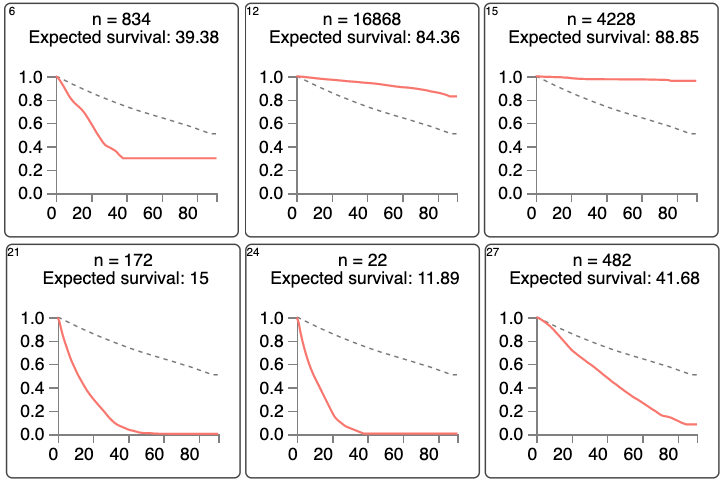}
  \caption{Survival curves from short-term Optimal Survival Tree for Nodes 6, 12, 15, 21, 24 and 27.}
  \label{fig:model-4-curves}
\end{figure}

By examining the Kaplan-Meier curve in each leaf, we see that the survival tree is successfully discriminating between failing and healthy leaf nodes on this short time scale. Figure \ref{fig:model-4-curves} shows the curves for selected leaves. We see that certain leaf nodes like Nodes 21 and 24 are comprised of extremely unhealthy drives, with near-certain failure within 90 days, while Nodes 12 and 15 contain seemingly healthy hard drives with almost no risk of failure within the 90-day window.

The Kaplan-Meier curves in the leaves are also a powerful tool for telling us more about the drive behavior than we can get from the expected survival time alone. Nodes 6 and 27 have very similar expected survival times (39.4 and 41.7 days, respectively), yet the curves show they exhibit distinctly different behaviors over time. It appears that drives in Node 6 tend to either fail within 40 days or not fail at all during the 90-day window. However, Node 27's drives are failing at a rate that is roughly constant over time. We can try to understand the differences in behavior by analyzing the conditions that lead to these leaves:

\begin{itemize}
  \item Node 6 is characterized by a high seek error rate (low values for normalized and high values for raw SMART metric 7), which is symptomatic of a partial failure in the mechanical positioning system: we might imagine that such a partial failure could be lethal in certain cases and surmountable in others, leading to the two classes of behavior seen in this node.
  \item Node 27 contains drives with relatively high values for raw SMART metric 5, which would normally be indicative of impending failure. However, these drives also have high values for normalized SMART metric 187, which might explain the steady pace of failure among these drives. This is in contrast to the extreme failure modes we observed in Nodes 21 and 24, which have anomalous values for both metrics 5 and 187.
\end{itemize}

In particular, comparing the results for Node 27 against those for Nodes 21 and 24 demonstrate the power of a non-linear model like Optimal Survival Trees. Individually, metrics 5 and 187 seem to be correlated with impending short-term failure, but the non-linearity of the tree uncovers that drives with abnormal values for metric 5 but normal values for 187 are in fact significantly healthier than those with abnormal values for both metrics.

\subsection{Short-term health as a classification problem} \label{oct}
Approaching the question of short-term health with survival analysis allows us to model the health of the drive over the entire 90-day time window, but sometimes we might want to focus on a specific operational concern, such as predicting whether drives are likely to fail within the next 30 days, to plan enough time for replacements. The survival analysis can be used to answer these questions, but because they model the health of the drive over the entire time window they may be less precise than a classification model trained specifically with this single focus.

To investigate this, we will pose the task of predicting drive failures within 30 days as a classification problem, and use Optimal Classification Trees to generate an interpretable predictive model for the probability of failure. Our target variable for the classification problem is whether or not the hard drive failed in the 30 days following the date the SMART metrics were recorded. We used one full year worth of data (Q2 2019 to Q1 2020, inclusive), and separated the data into training and testing based on the hard drives' serial numbers so that no drive had observations appearing in both the training and testing sets.

We show the trained Optimal Classification Tree in Figure \ref{fig:model-3-tree} (again, an interactive visualization with additional details is available in the supplementary materials). The probability of failure within 30 days as predicted by the model is displayed in each leaf node of the tree, with a darker color shading indicating higher probability of failure.

\begin{figure}
  \centering
  \includegraphics[width=\textwidth]{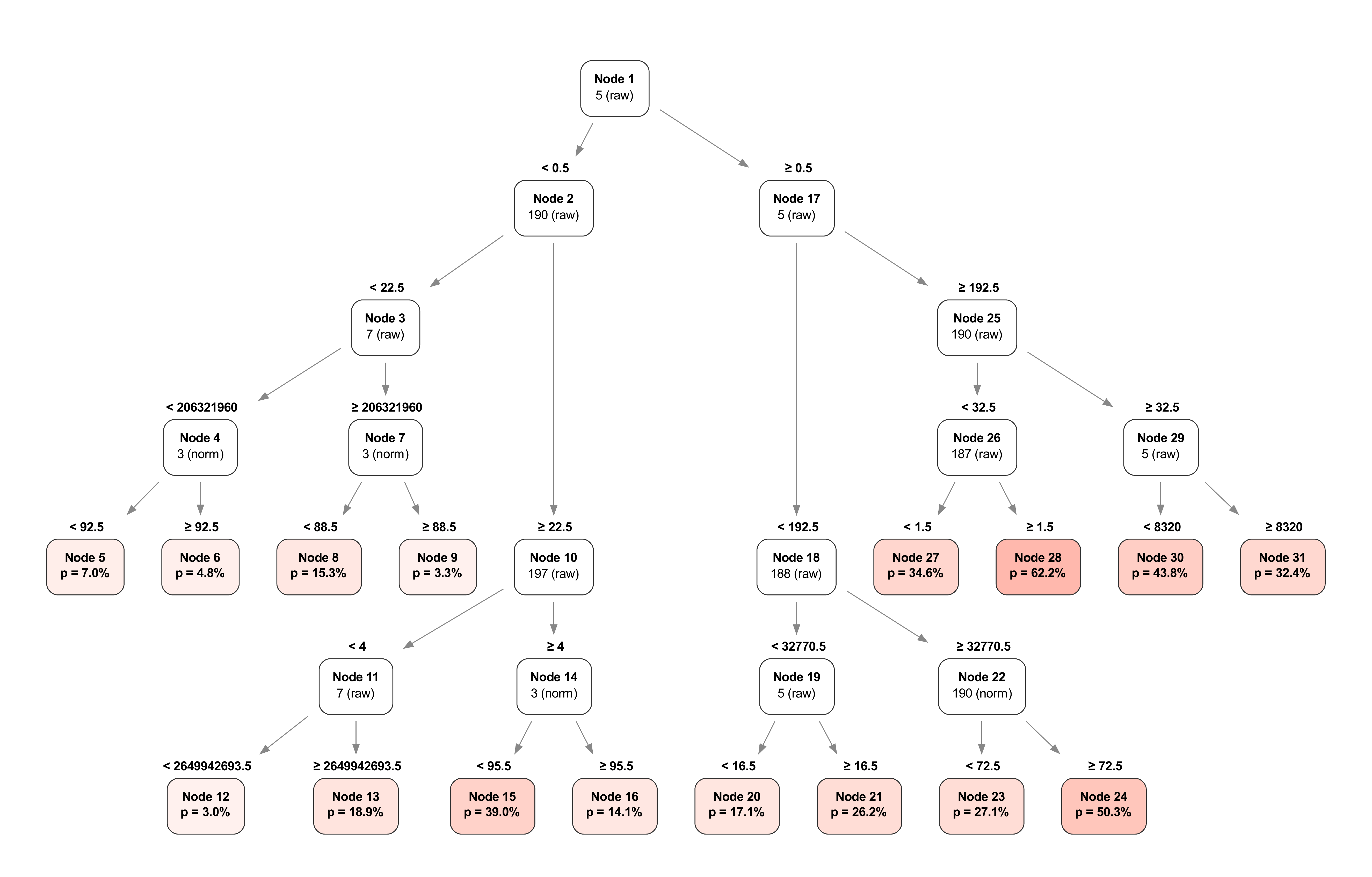}
  \caption{Optimal Classification Tree for predicting failure within 30 days. The variable in the split nodes refer to the splitting SMART variable (either as raw or normalized value). The probability in each leaf node is the predicted probability of failure, where darker shading indicates higher probability.}
  \label{fig:model-3-tree}
\end{figure}

\begin{figure}
  \centering
  \includegraphics[scale=0.65]{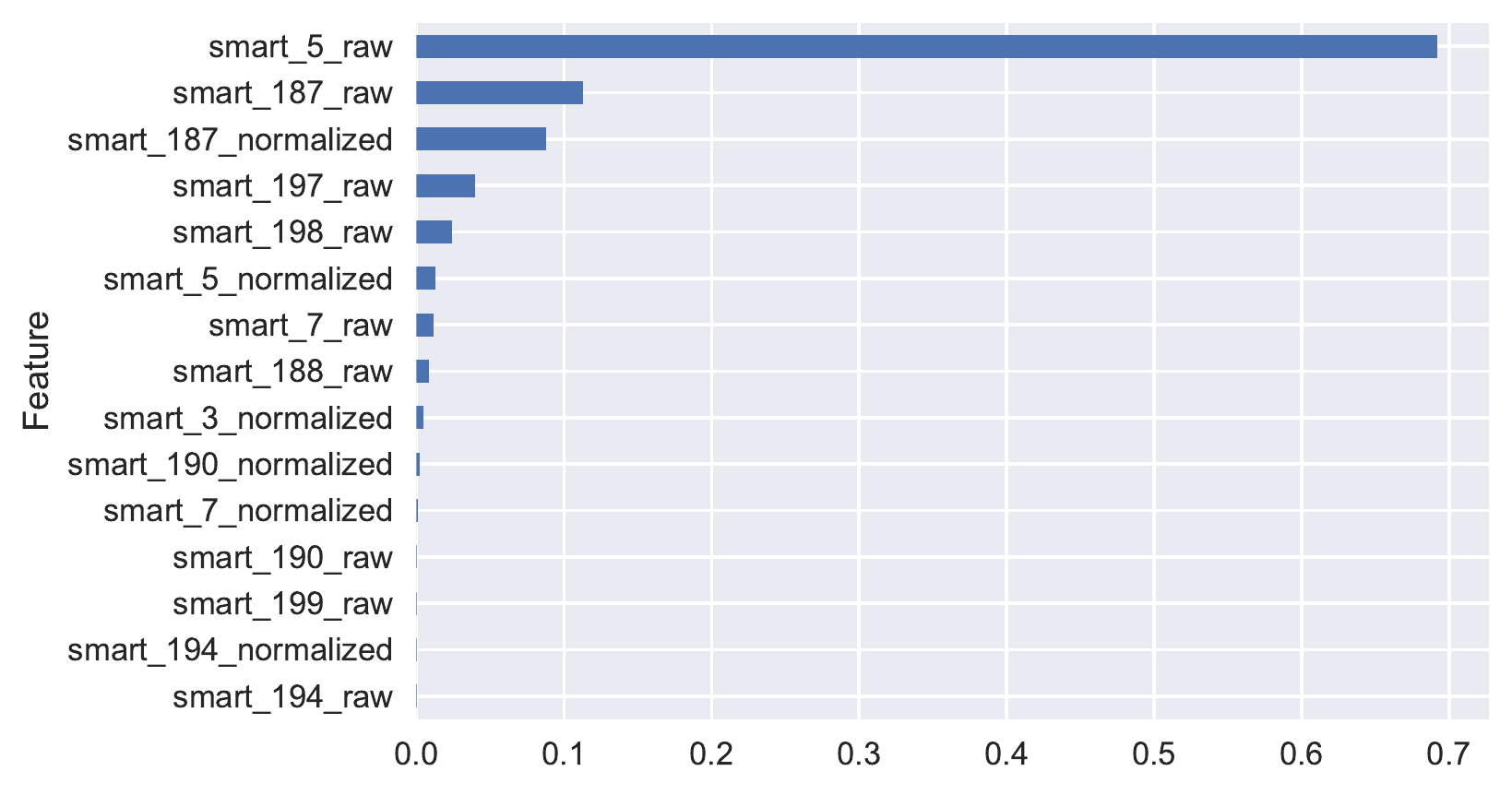}
  \caption{Variable importance for Optimal Classification Tree predicting failure within 30 days}
  \label{fig:model-3-importance}
\end{figure}

Inspecting the tree, we see that the structure appears similar to the survival tree presented in Section \ref{ost}, and similar variables are used in the splits. Figure \ref{fig:model-3-importance} shows the variable importance scores for this model, and indeed SMART metric 5 is again the most important, followed by 187. The probabilities of failure in each leaf range from 3.0\% to 62.2\%, indicating that the tree can indeed distinguish between the healthy drives and those with impending failure.

\subsection{Comparison of survival and classification approaches}

We have considered the short-term health from both survival and classification perspectives. The Optimal Classification Tree makes predictions of failure probability within 30 days, while the the Optimal Survival Tree allows us to make probabilistic predictions at any given time in 90-day window. Because the curve is monotone, the predicted probability of survival is always lower with longer evaluation time, consistent with intuition.

When assessing the quality of the predictions, we will report the following measures:
\begin{itemize}
    \item AUC is the area under the ROC (Receiver Operating Characteristic) curve, and is a holistic measure of the ability of the model to discriminate between failing and healthy drives, where 0.5 corresponds to random guessing and 1 indicates perfect predictions.
    \item Accuracy is the proportion of drives that are labeled correctly.
    \item Sensitivity, or true positive rate, is the percentage of failing machines which are correctly identified as failing: we want it as high as possible, as we would like to isolate failing machines for maintenance.
    \item The false alarm rate is the percentage of machines that are predicted to fail but do not actually fail: this should be as low as possible to minimize the number of drives falsely identified as being at risk of failure.
\end{itemize}

\begin{figure}
  \centering
  \includegraphics[scale=0.5]{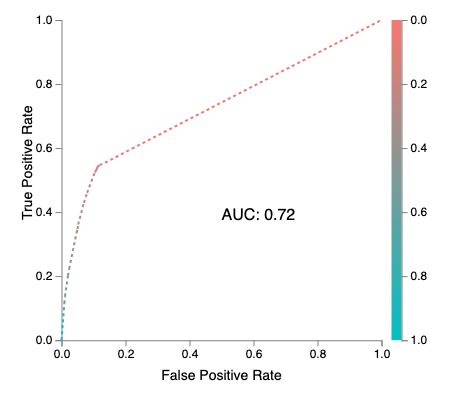}
  \caption{ROC curve for Optimal Classification Tree predicting failure within 30 days}
  \label{fig:model-3-roc}
\end{figure}

Accuracy, sensitivity and the false alarm rate all depend on the selection of a threshold for the probability of failure: if the predicted probability is above this threshold then the drive is predicted to fail. If the threshold is too low, then we will have high sensitivity but also a high false alarm rate, and conversely, if the threshold is too high, we will have a low number of false positives but also very low specificity. Figure \ref{fig:model-3-roc} shows the ROC curve which visualizes the trade-off between sensitivity and false alarm rate as the threshold is varied. To maximize the sensitivity while minimizing the false alarm rate, we elected to have the model operate at the inflection point of the curve, corresponding to a threshold of 5\%. In simple terms, if a machine's predicted probability of failing in the next 30 days is higher than 5\%, it is predicted as a failure.

Table \ref{tab:model-2-performance} shows the performance of the performance of the OCT (predicting failure within 30 days), and the OST when predicting failure within 30, 60, and 90 days. For each, we evaluate the AUC, accuracy, sensitivity, and false alarm rates, using the threshold of 5\% where required.

\begin{table}
    \centering
    \begin{tabular}{ l c c c c }
    \toprule
     \multirow{2}{*}{\textbf{Metric}} & \textbf{OCT} & \multicolumn{3}{c}{\textbf{OST}} \\
     \cmidrule(lr){2-2} \cmidrule(lr){3-5}
     & \textbf{30 day} &\textbf{30 day} & \textbf{60 day} & \textbf{90 day}\\
    \midrule
     \textbf{AUC}        & 0.724 & 0.692 & 0.664 & 0.701 \\
     \textbf{Accuracy}   & 0.861 & 0.853 & 0.839 & 0.833 \\
     \textbf{Sensitivity} & 0.547 & 0.520 & 0.463 & 0.450    \\
     \textbf{False Alarm Rate} & 0.119 & 0.126 & 0.116 & 0.089   \\
     \bottomrule
    \end{tabular}
    \caption{Performance results for OCT and OST at various evaluation times. Accuracy, sensitivity, and false alarm rate are evaluated under a threshold of 0.05.}\label{tab:model-2-performance}
\end{table}

We see that the performance of OST across all three time windows is similar, with sensitivity of roughly 45--50\% and false alarm rate around 10\%, indicating that we identify roughly half of the failing drives while only having a 10\% false positive rate. Comparing to the results of the OCT with the OST at 30 days, we see that the OCT has a small advantage in all metrics. This confirms our intuition that by directly formulating the problem as a classification problem for a specific time horizon of interest we can achieve better performance. In contrast, the survival tree is able to make predictions for any time horizon and guarantees monotonicity of these predictions over time, with a slight impact on performance compared to the classification approach.

\section{Conducting analysis with limited data} \label{limited-data}

In the last two sections, we have seen that extensive granular data representing the behavior of machines allows us to both assess the overall long-term health of such machines and predict which machines will fail over a shorter timeframe. However, it is not always the case that such a wealth of historical data is available for this modeling. In particular, collection of this kind of sensor data has often only begun recently, and in this situation a common question is \emph{how much data is required for an insightful analysis?} In this section, we will see that both analyses can still be conducted with data from a shorter time frame.

\subsection{Long-term health}

In Section \ref{long-term}, we analyzed the behavior of hard drives using a three-year time window, which covered the entire lifespan of nearly all the drives considered. To examine what is possible under a more data-limited scenario, here repeat our analysis using only a single year of data (Q1 2019 through Q1 2020, inclusive).

In the three-year case, we had sufficient failure data to restrict our analysis to hard drives that failed as we had sufficient data to observe their entire lifecycle. However, with only one year of data, the time window is too short to observe the entire lifespan for most failing drives. For this reason, we utilize all hard drives in the year of data for our analysis, and treat the drives that did not fail as censored data, as in Section \ref{ost}.

We show the trained survival tree in Figure \ref{fig:model-2-tree} (with the full interactive tree available in the supplementary materials). We see many similarities in structure between this tree and the one trained with full data from Section \ref{long-term}. For instance, the leaf with the lowest expected survival time in this tree is Node 5, characterized by a high value for raw SMART metric 5, and lower-than-baseline value for normalized SMART metrics 3 and 187. This is similar to the leaf with lowest expected survival in the original tree (Figure~\ref{fig:model-1-tree}), which is Node 18 and is characterized by a high raw value for metric 5 and 197, and deviations from baseline in metrics 3 and 187. Despite the shorter timeframe for the data, the tree is still able to discover similar modes of accelerated failure.

\begin{figure}
  \centering
  \includegraphics[ width=\textwidth]{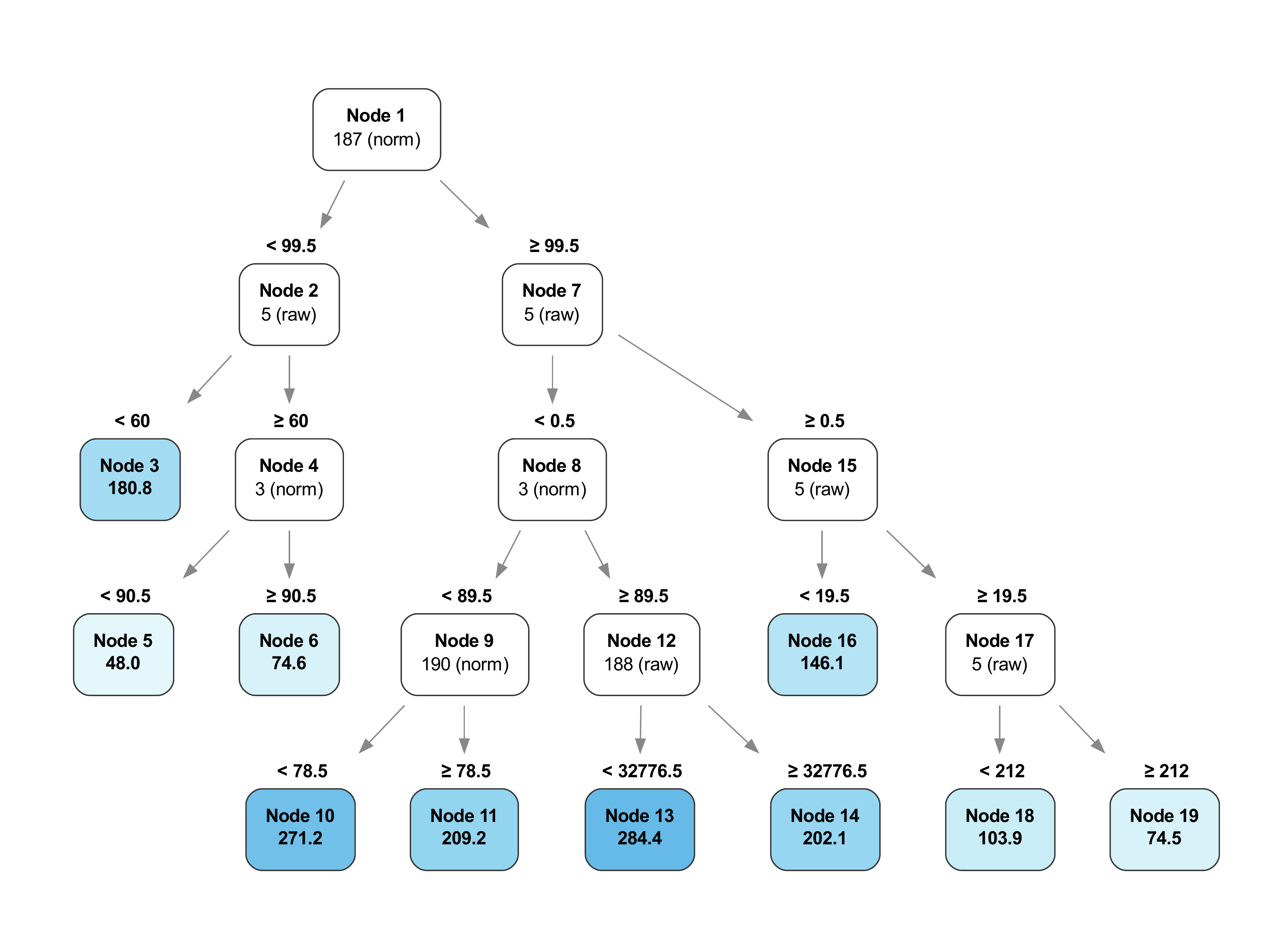}
  \caption{Optimal Survival Tree trained with limited data}
  \label{fig:model-2-tree}
\end{figure}

We also see that the tree includes a split on metric 190 (Temperature Difference). Recall that the univariate analysis conducted by Google concluded that this temperature was not predictive of impending hard drive failure. However, the tree uses it to fine-tune its prediction after analyzing metrics 3, 5, and 187, demonstrating the power and flexibility of such a non-linear model over simpler approaches.

Figure~\ref{fig:model-4-importance} shows the variable importance for the tree trained on limited data. We see that metric 5 is by far the most important, whereas metric 3 is much less important than it was for the original tree. This makes sense as metric 3 was used heavily on the left side of the tree (Figure~\ref{fig:model-1-tree}) to refine the survival estimates for the health drives with expected survival times between over 6 months. Given that the new tree is only exposed to a single year's worth of data, it makes sense that it cannot learn much about the longer-term survival in this way.

\begin{figure}
  \centering
  \includegraphics[scale=0.65]{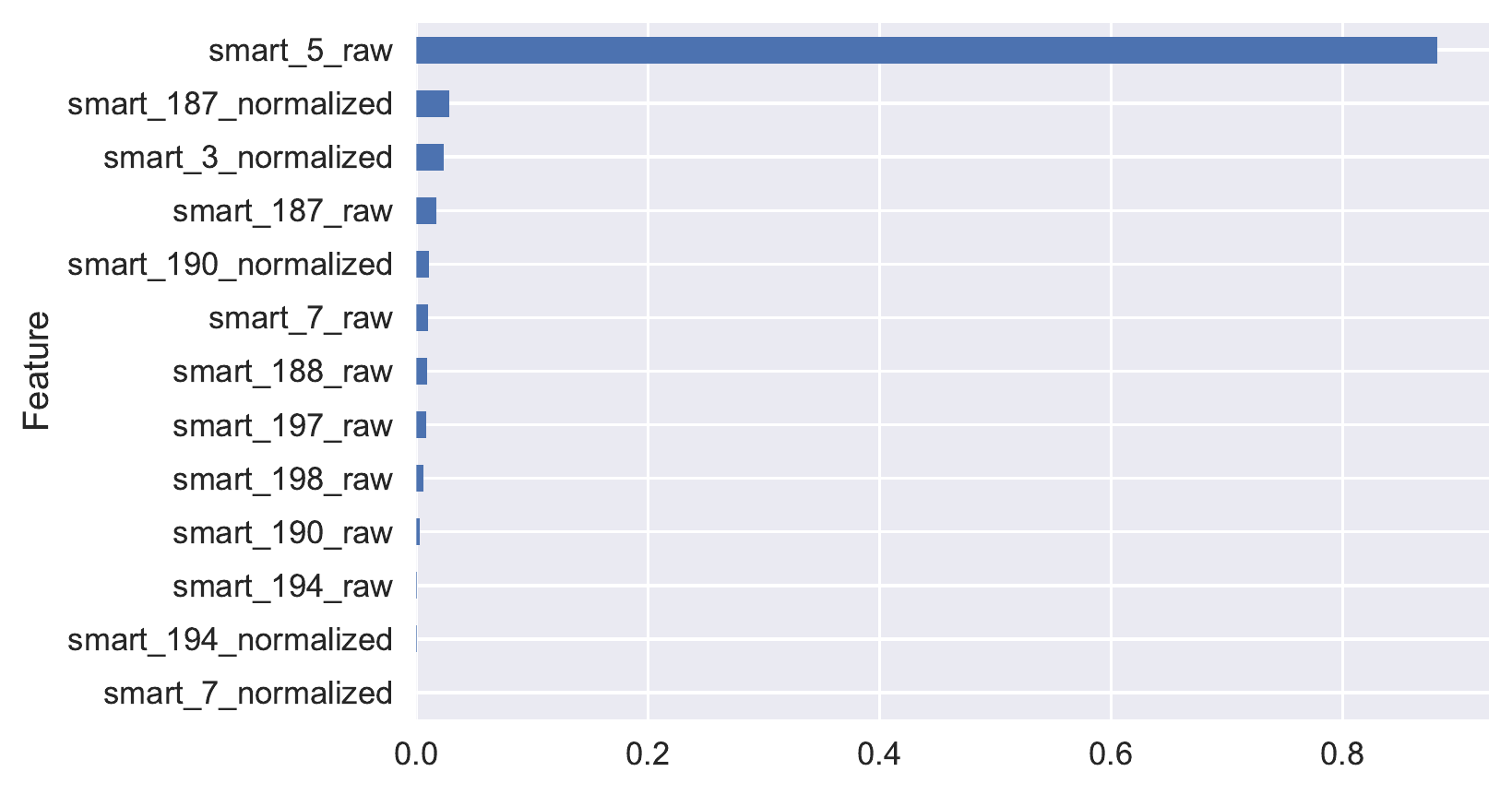}
  \caption{Variable importance for Optimal Survival Tree with limited data}
  \label{fig:model-2-importance}
\end{figure}

Overall, with only one year worth of data we see that it is still possible to recover useful insights about how the various SMART metrics indicate the overall health of a drive.

\subsection{Short-term health}

Analyzing the short-term health of drives requires no changes from before, provided we have data longer than the window of interest. The survival analysis in Section~\ref{ost} was already conducted using just a single quarter of data. The classification approach of~\ref{oct} used one year of data to run a thorough training and testing comparison analysis, but we could just as easily have obtained similar results with less data, provided that we observe a sufficient number of failures in line with our time window. Our intuition is that in this case as little as 1--2 quarters of data would be sufficient to predict 30-day failures.

\section{Conclusion}\label{conclusion}

In this paper, we showed that interpretable machine learning algorithms can be used to both assess the long-term health of hard drives, and to predict short-term failure. In addition, the approach can be adapted to limited data scenarios without significantly affecting the quality of the models.

The Optimal Classification Tree and Optimal Survival Tree models that we use bring a number of advantages to predictive maintenance over correlation analyses and classical predictive maintenance machine learning techniques:

\begin{itemize}
    \item \textbf{Detecting interpretable paths to failure}\\
    Optimal Trees can automatically identify both paths leading to accelerated failure as well as paths indicating healthy behaviors. Each leaf of the tree defines a cohort of hard drives with similar survival outcomes based on a series of conditions on their current SMART metric values, providing a discrete, easy-to-understand description that can be validated against expert knowledge. In this case, they uncovered interactions between several SMART metrics simultaneously in addition to confirming findings from existing univariate correlation analysis.

    \item \textbf{Tailored model to address each question} \\
    We presented many model variants each targeted at different questions. In Section~\ref{long-term}, we used OST to understand long-term health. In Section~\ref{ost}, we again used OST but for the purposes of characterizing short-term health. Finally, in Section~\ref{oct}, we used OCT to predict occurence of failure within a specific time window. In contrast to the one-size-fits-all style of previous analysis, this flexibility allows us to construct models that are targeted towards answering specific questions. These interpretable models can uncover simple-yet-powerful insights allowing data-oriented people and domain experts to have a common communication ground.

    \item \textbf{Applicable with limited data}
    If measurement data is scarce or is just starting to be collected, the approaches we utilized can still be applied to whatever limited data is present, and still generate useful and predictive models. In particular, the ability of Optimal Survival Trees to deal with censored data permits it to exploit any data that is available.
\end{itemize}

The analysis we conducted only considered a single model of drive, and most likely could be strengthened through approaches like engineering new derivative features based on changes in the SMART metrics over time. Nevertheless, the approach we present is a powerful way to derive useful insights and strong predictive performance by applying interpretable and non-linear methods to the problem of predictive maintenance.

\bibliography{ref}

\end{document}